# Personal Identification from Lip-Print Features using a Statistical Model

Saptarshi Bhattacharjee
Department of Information Technology
Institute of Engineering & Management, India

S Arunkumar
Department of Information Technology
Institute of Engineering & Management, India

Samir Kumar Bandyopadhyay
Professor
West Bengal University of Technology, India

## ABSTRACT

This paper presents a novel approach towards identification of human beings from the statistical analysis of their lip prints. Lip features are extracted by studying the spatial orientations of the grooves present in lip prints of individuals using standard edge detection techniques. Horizontal, vertical and diagonal groove features are analysed using connected-component analysis to generate the region-specific edge datasets. Comparison between test and reference sample datasets against a threshold value to define a match yield satisfactory results. FAR, FRR and ROC metrics have been used to gauge the performance of the algorithm for real-world deployment in *unimodal* and multimodal *biometric* verification systems.

## 1. INTRODUCTION

Biometrics-based authentication techniques have gained much importance in recent times. The main idea behind this approach is to identify human beings uniquely from their inherent physical traits. Identification from biometric parameters eradicates the problems associated with traditional methods of human identification. Human beings are identified from their physical features and not by some external object which they have to present for the process. Near accurate results are obtained as it is difficult to duplicate one's personal features.

Various well known methods have already been implemented in human identification (retina, iris, fingerprint, face etc.) [6, 7,8,9,10]. Although widespread progress has been made in this respect, it has been observed that even established biometric modalities fail to give accurate results in all real-life scenarios. Thus, novel biometric modalities are being researched on which can be used for identification effectively in real-world environment [15, 16].

In this paper we present forward a novel approach towards identification of human beings from the statistical analysis of their lip prints. Lip based identification approaches might not give results comparable with face of fingerprint verification techniques, yet these emerging modalities must be explored to increase efficiency in hybrid identification systems where more than one modality can be used to improve efficiency.

Lip prints can also be a basis for crime detection. It is used to find the situation on the basis of evidence surrounding the crime spot for identifying number of people involved, their nature, sex as well as type of crime committed during the event. Research studies and information regarding the use of lip prints as evidence in personal identification and criminal investigation are very much necessary.

In our study, two approaches have been demonstrated, namely '**Fast Match**' and '**Accurate Match**' for biometric authentication purposes. While the former refers to a fast and simple algorithm, the latter concerns a steady and more accurate one. Localized features are considered in the 'Accurate Match' method.

## 2. REVIEW WORKS

Lip print characteristics have been widely used in forensics by experts and by the law for human identification. While examining human lips characteristics the anatomical patterns on the lips are taken into account. Studies have shown that the grooves in the human lips are unique to each person, and hence can be used in human identification.

Although the study of Chieloscopy has gained much prominence in recent times, the idea was proposed in 1968 by Yasuo Tsuchihasi and Kazuo Suzuki at Tokyo University [1, 2]. They studied the lip prints of people of all ages and concluded that lip characteristics are unique and stable for a human being. Much recently, it has been studied that lip prints can also be used to determine the gender of a human being [5].

The pioneer of Chieloscopy, Professor J. Kasprzak, analysed 23 unique lip patterns [3-4] for finding features of human beings. Such patterns (lines, bifurcations, bridges, pentagons, dots, lakes, crossings, triangles etc.) are very similar to fingerprint, iris or palm print patterns. The statistical characteristics features extracted from the lip prints also account for unique identification.

Michal Choras has re-affirmed the belief in his recent studies [13, 14] that the lip can be used as a primary biometric modality for successful identification purposes. He has shown that geometrical analysis of the anatomical parameters of the human lip can be monitored for successful identification. Lukasz Smacki has also done significant research studying the groove patterns in the human lips for personal identification [11]. He has also proposed a method of lip print identification using DTW algorithm [12].

## 3. PROPOSED ALGORITHM

In our study, we propose two scanner based lip print authentication algorithms for successful biometric validation - The Fast Match algorithm and the Accurate match algorithm. The computational complexity of the Fast Match algorithm is superior over the Accurate Match algorithm, but the latter promises to yield more accurate matches in cases where the lip prints have a significant level of noise or are of low quality. The flowchart of the proposed algorithm is depicted in Fig 1.

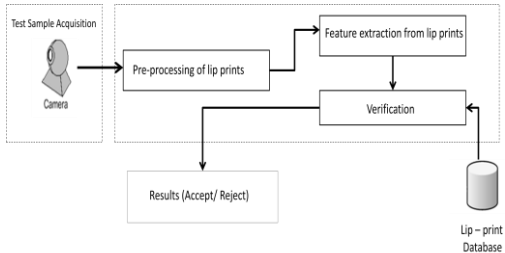

**Fig 1: The proposed algorithm modules**

In Fast Match, the training sample ($T_S$) is subjected to smoothing using a Gaussian Filter to generate the smoothed image ($T_F$). Prominent grooves in $T_F$ detected using the Canny edge detector to obtain the edge set ($T_C$) to be used for the verification phase. Sobel edge detection is a standard procedure to detect the vertical, horizontal and diagonal edges in an image using Sobel masks for vertical, horizontal and diagonal edge detection (Fig 2).

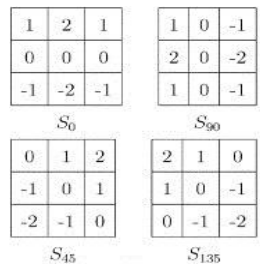

**Fig 2: The mask matrices of Sobel operator**

In Accurate match algorithm, the test sample ($T_S$) is further sub-divided into 4 blocks (Fig 7). $T_S$ is subjected to smoothing using a Gaussian Filter to generate the smoothed image ($T_F$). Prominent grooves in $T_F$ detected using the Canny edge detector to obtain the edge set ($T_C$) to be used for the verification phase. Sobel edge detection is a standard procedure to detect the vertical, horizontal and diagonal edges in an image using Sobel masks for vertical, horizontal and diagonal edge detection (Fig 2).

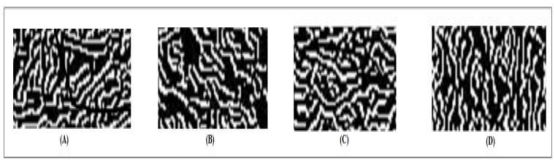

**Fig 3: Figure enunciating the various types of grooves detected by applying Sobel operator. Fig(A), Fig(B), Fig(C) and Fig(D) corresponding to $D_{1sobel}$, $D_{2sobel}$, $H_{sobel}$ and $V_{sobel}$ respectively.**

### 3.1. DATABASE

We have used the lip print database of Biometric Research Centre, University of Silesia at Katowice, Poland for our verification purposes [17].

In order to create the database, lip prints were collected from five individuals. For each person, four lip prints were stored. A procedure, very similar to those used in forensic labs for procuring comparative lip imprints was followed.

- Person applies a little bit of moisturizing cream to the red area of lip.
- After about two minutes, cheiloscopic roller with white sheet attached is pressed gently against the person's lips.
- Then the trace (print) left on white sheet is deciphered using black magnetic powder and magnetic applicator.
- Finally the white sheet containing the lip print was attached to a cheiloscopic card with some additional information.

After all the lip prints had been collected, the images were digitised. They were scanned at 600dpi resolution and were saved as bitmap images of 300 dpi.

The lip prints captured are bifurcated into Upper and Lower halves. Thus for each individual a database is created comprising of both the images of upper lip imprint as well as the lower one.

### 3.2. PRE-PROCESSING

The image obtained is converted into gray scale. Since the print is taken on a white paper, the background gets pregnant with noise (Fig 4) as an artefact. Thus the background accompanies uneven intensity variations. To abate the above aberration, clustering is done around the pixels with maximum and minimum intensity value to dichotomise the image into background and lip-imprint pixels.

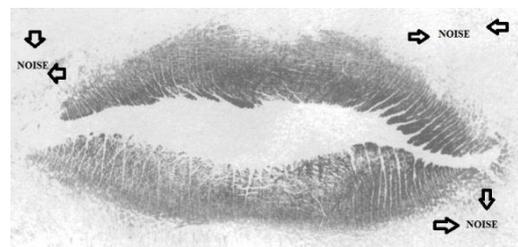

**Fig 4: Figure showing the obtained lip print along with the noise present in its background**

### 3.3. FEATURE EXTRACTION

Canny edge detection is used to enunciate all the groves present in the image. The cardinality of the edge set (after applying canny edge detection) is stored (say as $N_{canny}$). Now the feature extraction process bifurcates into two different approaches for 'Fast Match' and 'Accurate Match'.

### 3.3.1. FAST MATCH

In Fast Match, Sobel operator is used to compute the edge sets $E_H$, $E_V$, $E_{D1}$ and $E_{D2}$ representing the horizontal, vertical and diagonal grooves in I (Fig 3). Since each grove represents a connected component, the number of connected component present in each of $E_H$, $E_V$, $E_{D1}$ and $E_{D2}$ is stored as $H_{sobel}$, $V_{sobel}$, $D_{1sobel}$ and $D_{2sobel}$ respectively.

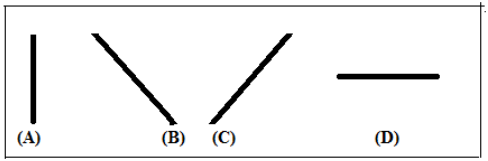

**Fig 5: Representation of four types of grooves: Fig (A) for 90 degree Fig (B) for 135 degree Fig(C) for 45 degree Fig (D) for 180 degree**

| 180degrees | Horizontal |
|---|---|
| 90degrees | Vertical |
| 45degrees | Diagonal1 |
| 135degrees | Diagonal2 |

**Fig 6: Figure showing the representation of the defined feature vector**

Edge set $E_C$ is extracted from $I$ using Canny Edge Detector and the cardinality of the set $E_C$ (denoted as $N_{canny}$) is used to normalize the data obtained from Sobel edge detection analysis in order to improve the efficiency.

Feature vector $F$ is defined comprising of elements $H_{norm}$, $V_{norm}$, $D_{1\,norm}$ and $D_{2\,norm}$, where

$$H_{norm} = \frac{H_{sobel}}{N_{canny}} \qquad V_{norm} = \frac{V_{sobel}}{N_{canny}}$$

$$D_{1\,norm} = \frac{D_{1\,sobel}}{N_{canny}} \qquad D_{2\,norm} = \frac{D_{2\,sobel}}{N_{canny}}$$

This feature vector is calculated for both the upper and lower lip imprint separately. Thus a feature vector is obtained having a total of eight elements (Feature vector of upper lip imprint + Feature vector of lower lip imprint) as
$F_{vector} = [F_{(upper)} \quad F_{(lower)}]$, where dimensions of $F_{(upper)}$ and $F_{(lower)}$ is 1x4.

### 3.3.2. ACCURATE MATCH
Accurate Match concentrates on extraction of lip features and subsequent analysis after localization in the original image $I$. The original image $I$ of dimensions [m, n] is resized to [m+k, n+k'] such that

$$\begin{cases} m + k = 4c \\ n + k' = 4c' \end{cases}, for\ some\ k, k', c, c' \in Z$$

Then, $I$ is split into 2x2 sub blocks $I_{11}$, $I_{12}$, $I_{21}$ and $I_{22}$ (Fig 7). The procedure followed after that is similar to that of 'Fast Match'. Feature vector is calculated for each block, represented by $F_{11}$, $F_{12}$, $F_{21}$ and $F_{22}$. The feature matrix is defined from the feature vectors for both the upper lip and the lower lip, so that the feature matrix ($F_{matrix}$) comprises of 4x(4+4) elements. The only difference from 'Fast Match' approach is that the feature matrix does not undergo any normalisation, since its effect on the efficiency is quite negligible (experimental observation). Feature vector $F_{(i,j)}$ comprises of elements $H_{(i,j)sobel}$, $V_{(i,j)sobel}$, $D_{1(i,j)sobel}$ and $D_{2(i,j)sobel}$ respectively, is defined as

$$F_{(i,j)} = \begin{bmatrix} H_{(i,j)_{sobel}} & V_{(i,j)_{sobel}} & D_{1(i,j)_{sobel}} & D_{2(i,j)_{sobel}} \end{bmatrix},$$
where $i, j \leq 2$

and $\qquad F_{matrix} = \begin{bmatrix} F_{11} \\ F_{12} \\ F_{21} \\ F_{22} \end{bmatrix}$

This feature matrix is calculated for both the upper and lower lip imprint separately. Thus the final feature matrix is of dimension 4x8, where we have the features for both the upper and the lower lip as

$F = [F_{matrix(upper)} \quad F_{matrix(lower)}]$ where dimensions of $\mathbf{F_{matrix(upper)}}$ and $\mathbf{F_{matrix(lower)}}$ is 4x4.

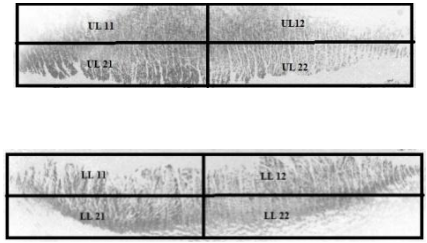

**Fig 7: Figure depicting the division of the given image in blocks**

## 4. VERIFICATION
### 4.1. FAST MATCH
The feature vectors from the test samples ($F_T$) and reference samples ($F_R$) were compared to detect a match. Euclidean distance is a statistical measure to compute the distance between two vectors for matching purposes. The Euclidean distance (**D**) between $\mathbf{F_T}$ and $\mathbf{F_R}$ is computed and compared with the threshold acceptance value ($T_v$) to decide an accept or reject.

$$D_{fast} = \sqrt{\sum_{i=1}^{4}(Ft(i) - Fr(i))^2}$$

$$D_{fast} = \begin{cases} \leq Tv \rightarrow Accept \\ > Tv \rightarrow Reject \end{cases}$$

### 4.2. ACCURATE MATCH
The feature matrices from the test samples ($F_T$) and reference samples ($F_R$) were compared to detect a match. Euclidean distance is a statistical measure to compute the distance between two matrices for matching purposes. The Euclidean distance (**D**) between $\mathbf{F_T}$ and $\mathbf{F_R}$ is computed and compared with the threshold acceptance value (**T**) to decide an accept or reject.

$$D_{Accurate} = \sqrt{\sum_{i=1}^{4}\sum_{j=1}^{4}(Ft(i,j) - Fr(i,j))^2}$$

$$D_{Accurate} = \begin{cases} \leq Tv \rightarrow Accept \\ > Tv \rightarrow Reject \end{cases}$$

## 5. EXPERIMENT RESULTS

The algorithm was tested on database comprising of 20 lip prints (5 sample lip prints were taken from four individuals). The first three samples of every individual was used for training ( i.e. to calculate the threshold of Euclidean distance). Threshold was defined as the maximum of the Euclidean distances obtained when the lip prints (first three) of the same individuals were matched. After defining the threshold, our algorithm was tested on the complete database. The results obtained are enunciated by the tables (Table 1 and Table 2) shown below.

**Table 1: Results obtained for 'Fast Match' Algorithm.**

| Sample Size | FAR (%) | FRR (%) | Efficiency(%) |
|---|---|---|---|
| 5 | 15.2188 | 2.0528 | 82.7284 |
| 10 | 12.4063 | 1.6790 | 85.9147 |
| 20 | 11.2500 | 0.8750 | 87.8750 |

**Table 2: Results obtained for 'Accurate Match' Algorithm.**

| Sample Size | FAR (%) | FRR (%) | Efficiency(%) |
|---|---|---|---|
| 5 | 7.0968 | 1.2145 | 91.6887 |
| 10 | 6.0938 | 0.6350 | 93.2712 |
| 20 | 3.4375 | 0.0825 | 96.4800 |

## 6. CONCLUSION AND FUTURE WORKS

It could be clearly learnt from the results displayed in Table 1 and Table 2 that the FRR is less than that of FAR. Moreover the FAR also considerably reduces as the sample size increases.

It could also be anticipated that the efficiency will increase further if the sample size increases (since the FAR will get considerably reduced).

Future Work is to increase the size of the feature vector (i.e. to define more directions like 25, 75,155 degrees etc) and study its results. Measures to reduce the FAR would also be considered in future.

The major application of this algorithm could be found in the implementation of a hybrid biometric system, where it could be used in conjunction with some other biometric modality to yield better accuracy.

Since this a novel approach towards lip print based biometric authentication using connected component analysis in conjunction with edge detection, no comparison can be provided with other works.

Since there is no public database of lip prints available for this kind f study, we had to test our results on a set of 40 prints only. We hope to test it in future for a large database with a significant diversity in the nature and origin of the lip prints.

## 7. ACKNOWLEDGEMENT

We would like to extend our gratitude to Prof. Lukasz Smacki [18] of the University of Silesia in Katowice (Poland), for providing us with the database of lip-prints of various individuals. Without his support, our advancement in this field of research would not have been possible.

## 8. REFERENCES

[1] Tsuchihasi, Y.: "Studies on Personal Identification by Means of Lip Prints" Forensic Science 3(3) (1974).

[2] Suzuki K., Tsuchihashi Y.: "personal identification by means of lip prints", J.Forensic Med.17:52-57, 1970.

[3] Kasprzak J, Leczynska B (2001) Chieloscopy. "Human identification on the basis of lip Prints" (in Polish). CLK KGP Press,Warsaw, 2001

[4] Kasprzak J. Possibilities of cheiloscopy. Forensic Sci Int ,; 46: 145 – 151, 1990.

[5] Sonal, V., Nayak, C.D., Pagare, S.S.: "Study of Lip-Prints as Aid for Sex Determination", Medico-Legal Update 5(3) (2005).

[6] CC Han, HL Cheng, CL Lin, KC Fan. "Personal authentication using palm-print features" - Pattern Recognition, 2003 – Elsevier

[7] Yunhong Wang, Tieniu Tan, Anil K. Jain. "Combining Face and Iris Biometrics for Identity Verification" - Lecture Notes in Computer Science, Springer, 2003. Volume 2688

[8] S Lim, K Lee, O Byeon, T Kim. "Efficient Iris Recognition through Improvement of Feature Vector and Classifier" - ETRI journal, 2001 - etrij.etri.re.kr

[9] Ribarić, Slobodan; Fratrić, Ivan. "A Biometric Verification System Based on the Fusion of Palmprint and Face Features" - Proceedings of the 4th International Symposium on Image and Signal Processing and Analysis, Zagreb, 2005.

[10] A.H. Mir, S. Rubab, Z. A. Jhat. "Biometrics Verification: A Literature Survey" - International Journal of Computing and ICT Research, Vol. 5, No.2, December 2011

[11] Łukasz Smacki, Piotr Porwik, Krzysztof Tomaszycki, Sylwia Kwarcińska. "The Lip Print Recognition using Hough Transform" - Journal of Medical Informatics & Technologies Vol. 14/2010

[12] Lukasz Smacki and Krzysztof Wrobel. "Lip Print Recognition Based on Mean Differences Similarity Measure" - Advances in Intelligent and Soft Computing, 2011, Volume 95/2011, Computer Recognition Systems 4, Springer

[13] Michał Choras´. "The lip as a biometric" - Pattern Analysis & Applications, 2010 – Springer

[14] Choras' M. , Emerging Methods of Biometrics Human Identification. In: Proc. of ICICIC 2007 - Kummamoto, Japan, IEEE CS Press, 2007

[15] Prabhakar S., Kittler J., Maltoni D., O'Gorman L., Tan T. , "Introduction to the Special Issue on Biometrics: Progress and Directions", IEEE Trans. on PAMI, vol. 29, no. 4, 513-516, 2007

[16] Goudelis G., Tefas A., Pitas I. , "On Emerging Biometric Technologies". In Proc. of COST 275 Biometrics on the Internet, 71-74, Hatfield UK, 2005

[17] Biometric Research Centre, University of Silesia, Katowice, Poland. http://www.biometrics.us.edu.pl/

[18] Prof Lukasz Smacki, , University of Silesia, Katowice, Poland. http://www.biometrics.us.edu.pl/about-us/people/lukasz-smacki

[18] Prof Lukasz Smacki, , University of Silesia, Katowice, Poland. http://www.biometrics.us.edu.pl/about-us/people/lukasz-smacki

[19] Smacki, Lip traces recognition based on lines pattern ; Journal of Medical Informatics & Technologies Vol. 15/2010, ISSN 1642-6037

[20] Olufemi Sunday Adeoye, A Survey of Emerging Biometric Technologies; International Journal of Computer Applications Vol 10  2010

[21] DOROZ R., PORWIK P., PARA T., WRÓBEL K., Dynamic signature recognition based on velocity changes of some features, International Journal of Biometrics, Vol. 1, No. 1, 2008, pp. 47–62.

[22] Sushil Chauhan, A.S. Arora, Amit Kaul ; A survey of emerging biometric modalities ; Proceedings of the International Conference and Exhibition on Biometrics Technology, Elsevier